\documentclass[conference]{IEEEtran}
\IEEEoverridecommandlockouts
\usepackage{cite}
\usepackage{amsmath,amssymb,amsfonts}
\usepackage{algorithmic}
\usepackage{graphicx}
\usepackage{textcomp}
\usepackage{xcolor}
\def\BibTeX{{\rm B\kern-.05em{\sc i\kern-.025em b}\kern-.08em
    T\kern-.1667em\lower.7ex\hbox{E}\kern-.125emX}}

\def\PsfigVersion{1.9}
\ifx\undefined\psfig\else\endinput\fi

\let\LaTeXAtSign=\@
\let\@=\relax
\edef\psfigRestoreAt{\catcode`\@=\number\catcode`@\relax}
\catcode`\@=11\relax
\newwrite\@unused
\def\ps@typeout#1{{\let\protect\string\immediate\write\@unused{#1}}}
\ps@typeout{psfig/tex \PsfigVersion}

\def\figurepath{./}

\def\@nnil{\@nil}
\def\@empty{}
\def\@psdonoop#1\@@#2#3{}
\def\@psdo#1:=#2\do#3{\edef\@psdotmp{#2}\ifx\@psdotmp\@empty \else
    \expandafter\@psdoloop#2,\@nil,\@nil\@@#1{#3}\fi}
\def\@psdoloop#1,#2,#3\@@#4#5{\def#4{#1}\ifx #4\@nnil \else
       #5\def#4{#2}\ifx #4\@nnil \else#5\@ipsdoloop #3\@@#4{#5}\fi\fi}
\def\@ipsdoloop#1,#2\@@#3#4{\def#3{#1}\ifx #3\@nnil 
       \let\@nextwhile=\@psdonoop \else
      #4\relax\let\@nextwhile=\@ipsdoloop\fi\@nextwhile#2\@@#3{#4}}
\def\@tpsdo#1:=#2\do#3{\xdef\@psdotmp{#2}\ifx\@psdotmp\@empty \else
    \@tpsdoloop#2\@nil\@nil\@@#1{#3}\fi}
\def\@tpsdoloop#1#2\@@#3#4{\def#3{#1}\ifx #3\@nnil 
       \let\@nextwhile=\@psdonoop \else
      #4\relax\let\@nextwhile=\@tpsdoloop\fi\@nextwhile#2\@@#3{#4}}
\ifx\undefined\fbox
\newdimen\fboxrule
\newdimen\fboxsep
\newdimen\ps@tempdima
\newbox\ps@tempboxa
\long\def\fbox#1{\leavevmode\setbox\ps@tempboxa\hbox{#1}\ps@tempdima\fboxrule
    \advance\ps@tempdima \fboxsep \advance\ps@tempdima \dp\ps@tempboxa
   \hbox{\lower \ps@tempdima\hbox
  {\vbox{\hrule height \fboxrule
          \hbox{\vrule width \fboxrule \hskip\fboxsep
          \vbox{\vskip\fboxsep \box\ps@tempboxa\vskip\fboxsep}\hskip 
                 \fboxsep\vrule width \fboxrule}
                 \hrule height \fboxrule}}}}
\fi
\newread\ps@stream
\newif\ifnot@eof       % continue looking for the bounding box?
\newif\if@noisy        % report what you're making?
\newif\if@atend        % %%BoundingBox: has (at end) specification
\newif\if@psfile       % does this look like a PostScript file?
{\catcode`\%=12\global\gdef\epsf@start{%!}}
\def\epsf@PS{PS}
\def\epsf@getbb#1{%
\openin\ps@stream=#1
\ifeof\ps@stream\ps@typeout{Error, File #1 not found}\else
   {\not@eoftrue \chardef\other=12
    \def\do##1{\catcode`##1=\other}\dospecials \catcode`\ =10
    \loop
       \if@psfile
	  \read\ps@stream to \epsf@fileline
       \else{
	  \obeyspaces
          \read\ps@stream to \epsf@tmp\global\let\epsf@fileline\epsf@tmp}
       \fi
       \ifeof\ps@stream\not@eoffalse\else
       \if@psfile\else
       \expandafter\epsf@test\epsf@fileline:. \\%
       \fi
          \expandafter\epsf@aux\epsf@fileline:. \\%
       \fi
   \ifnot@eof\repeat
   }\closein\ps@stream\fi}%
\long\def\epsf@test#1#2#3:#4\\{\def\epsf@testit{#1#2}
			\ifx\epsf@testit\epsf@start\else
\ps@typeout{Warning! File does not start with `\epsf@start'.  It may not be a PostScript file.}
			\fi
			\@psfiletrue} % don't test after 1st line
{\catcode`\%=12\global\let\epsf@percent=%\global\def\epsf@bblit{%BoundingBox}}
\long\def\epsf@aux#1#2:#3\\{\ifx#1\epsf@percent
   \def\epsf@testit{#2}\ifx\epsf@testit\epsf@bblit
	\@atendfalse
        \epsf@atend #3 . \\%
	\if@atend	
	   \if@verbose{
		\ps@typeout{psfig: found `(atend)'; continuing search}
	   }\fi
        \else
        \epsf@grab #3 . . . \\%
        \not@eoffalse
        \global\no@bbfalse
        \fi
   \fi\fi}%
\def\epsf@grab #1 #2 #3 #4 #5\\{%
   \global\def\epsf@llx{#1}\ifx\epsf@llx\empty
      \epsf@grab #2 #3 #4 #5 .\\\else
   \global\def\epsf@lly{#2}%
   \global\def\epsf@urx{#3}\global\def\epsf@ury{#4}\fi}%
\def\epsf@atendlit{(atend)} 
\def\epsf@atend #1 #2 #3\\{%
   \def\epsf@tmp{#1}\ifx\epsf@tmp\empty
      \epsf@atend #2 #3 .\\\else
   \ifx\epsf@tmp\epsf@atendlit\@atendtrue\fi\fi}

\chardef\psletter = 11 % won't conflict with \begin{letter} now...
\chardef\other = 12

\newif \ifdebug %%% turn me on to see TeX hard at work ...
\newif\ifc@mpute %%% don't need to compute some values
\c@mputetrue % but assume that we do

\let\then = \relax
\def\r@dian{pt }
\let\r@dians = \r@dian
\let\dimensionless@nit = \r@dian
\let\dimensionless@nits = \dimensionless@nit
\def\internal@nit{sp }
\let\internal@nits = \internal@nit
\newif\ifstillc@nverging
\def \Mess@ge #1{\ifdebug \then \message {#1} \fi}

{ %%% Things that need abnormal catcodes %%%
	\catcode `\@ = \psletter
	\gdef \nodimen {\expandafter \n@dimen \the \dimen}
	\gdef \term #1 #2 #3%
	       {\edef \t@ {\the #1}%%% freeze parameter 1 (count, by value)
		\edef \t@@ {\expandafter \n@dimen \the #2\r@dian}%
		\t@rm {\t@} {\t@@} {#3}%
	       }
	\gdef \t@rm #1 #2 #3%
	       {{%
		\count 0 = 0
		\dimen 0 = 1 \dimensionless@nit
		\dimen 2 = #2\relax
		\Mess@ge {Calculating term #1 of \nodimen 2}%
		\loop
		\ifnum	\count 0 < #1
		\then	\advance \count 0 by 1
			\Mess@ge {Iteration \the \count 0 \space}%
			\Multiply \dimen 0 by {\dimen 2}%
			\Mess@ge {After multiplication, term = \nodimen 0}%
			\Divide \dimen 0 by {\count 0}%
			\Mess@ge {After division, term = \nodimen 0}%
		\repeat
		\Mess@ge {Final value for term #1 of 
				\nodimen 2 \space is \nodimen 0}%
		\xdef \Term {#3 = \nodimen 0 \r@dians}%
		\aftergroup \Term
	       }}
	\catcode `\p = \other
	\catcode `\t = \other
	\gdef \n@dimen #1pt{#1} %%% throw away the ``pt''
}

\def \Divide #1by #2{\divide #1 by #2} %%% just a synonym

\def \Multiply #1by #2%%% allows division of a dimen by a dimen
       {{%%% should really freeze parameter 2 (dimen, passed by value)
	\count 0 = #1\relax
	\count 2 = #2\relax
	\count 4 = 65536
	\Mess@ge {Before scaling, count 0 = \the \count 0 \space and
			count 2 = \the \count 2}%
	\ifnum	\count 0 > 32767 %%% do our best to avoid overflow
	\then	\divide \count 0 by 4
		\divide \count 4 by 4
	\else	\ifnum	\count 0 < -32767
		\then	\divide \count 0 by 4
			\divide \count 4 by 4
		\else
		\fi
	\fi
	\ifnum	\count 2 > 32767 %%% while retaining reasonable accuracy
	\then	\divide \count 2 by 4
		\divide \count 4 by 4
	\else	\ifnum	\count 2 < -32767
		\then	\divide \count 2 by 4
			\divide \count 4 by 4
		\else
		\fi
	\fi
	\multiply \count 0 by \count 2
	\divide \count 0 by \count 4
	\xdef \product {#1 = \the \count 0 \internal@nits}%
	\aftergroup \product
       }}

\def\r@duce{\ifdim\dimen0 > 90\r@dian \then   % sin(x+90) = sin(180-x)
		\multiply\dimen0 by -1
		\advance\dimen0 by 180\r@dian
		\r@duce
	    \else \ifdim\dimen0 < -90\r@dian \then  % sin(-x) = sin(360+x)
		\advance\dimen0 by 360\r@dian
		\r@duce
		\fi
	    \fi}

\def\Sine#1%
       {{%
	\dimen 0 = #1 \r@dian
	\r@duce
	\ifdim\dimen0 = -90\r@dian \then
	   \dimen4 = -1\r@dian
	   \c@mputefalse
	\fi
	\ifdim\dimen0 = 90\r@dian \then
	   \dimen4 = 1\r@dian
	   \c@mputefalse
	\fi
	\ifdim\dimen0 = 0\r@dian \then
	   \dimen4 = 0\r@dian
	   \c@mputefalse
	\fi
	\ifc@mpute \then
		\divide\dimen0 by 180
		\dimen0=3.141592654\dimen0
		\dimen 2 = 3.1415926535897963\r@dian %%% a well-known constant
		\divide\dimen 2 by 2 %%% we only deal with -pi/2 : pi/2
		\Mess@ge {Sin: calculating Sin of \nodimen 0}%
		\count 0 = 1 %%% see power-series expansion for sine
		\dimen 2 = 1 \r@dian %%% ditto
		\dimen 4 = 0 \r@dian %%% ditto
		\loop
			\ifnum	\dimen 2 = 0 %%% then we've done
			\then	\stillc@nvergingfalse 
			\else	\stillc@nvergingtrue
			\fi
			\ifstillc@nverging %%% then calculate next term
			\then	\term {\count 0} {\dimen 0} {\dimen 2}%
				\advance \count 0 by 2
				\count 2 = \count 0
				\divide \count 2 by 2
				\ifodd	\count 2 %%% signs alternate
				\then	\advance \dimen 4 by \dimen 2
				\else	\advance \dimen 4 by -\dimen 2
				\fi
		\repeat
	\fi		
			\xdef \sine {\nodimen 4}%
       }}

\def\Cosine#1{\ifx\sine\UnDefined\edef\Savesine{\relax}\else
		             \edef\Savesine{\sine}\fi
	{\dimen0=#1\r@dian\advance\dimen0 by 90\r@dian
	 \Sine{\nodimen 0}
	 \xdef\cosine{\sine}
	 \xdef\sine{\Savesine}}}	      
\def\psdraft{
	\def\@psdraft{0}
}
\def\psfull{
	\def\@psdraft{100}
}

\psfull

\newif\if@scalefirst
\def\psscalefirst{\@scalefirsttrue}
\def\psrotatefirst{\@scalefirstfalse}
\psrotatefirst

\newif\if@draftbox
\def\psnodraftbox{
	\@draftboxfalse
}
\def\psdraftbox{
	\@draftboxtrue
}
\@draftboxtrue

\newif\if@prologfile
\newif\if@postlogfile
\def\pssilent{
	\@noisyfalse
}
\def\psnoisy{
	\@noisytrue
}
\psnoisy
\newif\if@bbllx
\newif\if@bblly
\newif\if@bburx
\newif\if@bbury
\newif\if@height
\newif\if@width
\newif\if@rheight
\newif\if@rwidth
\newif\if@angle
\newif\if@clip
\newif\if@verbose
\def\@p@@sclip#1{\@cliptrue}

\newif\if@decmpr

\def\@p@@sfigure#1{\def\@p@sfile{null}\def\@p@sbbfile{null}
	        \openin1=#1.bb
		\ifeof1\closein1
	        	\openin1=\figurepath#1.bb
			\ifeof1\closein1
			        \openin1=#1
				\ifeof1\closein1%
				       \openin1=\figurepath#1
					\ifeof1
					   \ps@typeout{Error, File #1 not found}
						\if@bbllx\if@bblly
				   		\if@bburx\if@bbury
			      				\def\@p@sfile{#1}%
			      				\def\@p@sbbfile{#1}%
							\@decmprfalse
				  	   	\fi\fi\fi\fi
					\else\closein1
				    		\def\@p@sfile{\figurepath#1}%
				    		\def\@p@sbbfile{\figurepath#1}%
						\@decmprfalse
	                       		\fi%
			 	\else\closein1%
					\def\@p@sfile{#1}
					\def\@p@sbbfile{#1}
					\@decmprfalse
			 	\fi
			\else
				\def\@p@sfile{\figurepath#1}
				\def\@p@sbbfile{\figurepath#1.bb}
				\@decmprtrue
			\fi
		\else
			\def\@p@sfile{#1}
			\def\@p@sbbfile{#1.bb}
			\@decmprtrue
		\fi}

\def\@p@@sfile#1{\@p@@sfigure{#1}}

\def\@p@@sbbllx#1{
		\@bbllxtrue
		\dimen100=#1
		\edef\@p@sbbllx{\number\dimen100}
}
\def\@p@@sbblly#1{
		\@bbllytrue
		\dimen100=#1
		\edef\@p@sbblly{\number\dimen100}
}
\def\@p@@sbburx#1{
		\@bburxtrue
		\dimen100=#1
		\edef\@p@sbburx{\number\dimen100}
}
\def\@p@@sbbury#1{
		\@bburytrue
		\dimen100=#1
		\edef\@p@sbbury{\number\dimen100}
}
\def\@p@@sheight#1{
		\@heighttrue
		\dimen100=#1
   		\edef\@p@sheight{\number\dimen100}
}
\def\@p@@swidth#1{
		\@widthtrue
		\dimen100=#1
		\edef\@p@swidth{\number\dimen100}
}
\def\@p@@srheight#1{
		\@rheighttrue
		\dimen100=#1
		\edef\@p@srheight{\number\dimen100}
}
\def\@p@@srwidth#1{
		\@rwidthtrue
		\dimen100=#1
		\edef\@p@srwidth{\number\dimen100}
}
\def\@p@@sangle#1{
		\@angletrue
		\edef\@p@sangle{#1} %\number\dimen100}
}
\def\@p@@ssilent#1{ 
		\@verbosefalse
}
\def\@p@@sprolog#1{\@prologfiletrue\def\@prologfileval{#1}}
\def\@p@@spostlog#1{\@postlogfiletrue\def\@postlogfileval{#1}}
\def\@cs@name#1{\csname #1\endcsname}
\def\@setparms#1=#2,{\@cs@name{@p@@s#1}{#2}}
\def\ps@init@parms{
		\@bbllxfalse \@bbllyfalse
		\@bburxfalse \@bburyfalse
		\@heightfalse \@widthfalse
		\@rheightfalse \@rwidthfalse
		\def\@p@sbbllx{}\def\@p@sbblly{}
		\def\@p@sbburx{}\def\@p@sbbury{}
		\def\@p@sheight{}\def\@p@swidth{}
		\def\@p@srheight{}\def\@p@srwidth{}
		\def\@p@sangle{0}
		\def\@p@sfile{} \def\@p@sbbfile{}
		\def\@p@scost{10}
		\def\@sc{}
		\@prologfilefalse
		\@postlogfilefalse
		\@clipfalse
		\if@noisy
			\@verbosetrue
		\else
			\@verbosefalse
		\fi
}
\def\parse@ps@parms#1{
	 	\@psdo\@psfiga:=#1\do
		   {\expandafter\@setparms\@psfiga,}}
\newif\ifno@bb
\def\bb@missing{
	\if@verbose{
		\ps@typeout{psfig: searching \@p@sbbfile \space  for bounding box}
	}\fi
	\no@bbtrue
	\epsf@getbb{\@p@sbbfile}
        \ifno@bb \else \bb@cull\epsf@llx\epsf@lly\epsf@urx\epsf@ury\fi
}	
\def\bb@cull#1#2#3#4{
	\dimen100=#1 bp\edef\@p@sbbllx{\number\dimen100}
	\dimen100=#2 bp\edef\@p@sbblly{\number\dimen100}
	\dimen100=#3 bp\edef\@p@sbburx{\number\dimen100}
	\dimen100=#4 bp\edef\@p@sbbury{\number\dimen100}
	\no@bbfalse
}
\newdimen\p@intvaluex
\newdimen\p@intvaluey
\def\rotate@#1#2{{\dimen0=#1 sp\dimen1=#2 sp
		  \global\p@intvaluex=\cosine\dimen0
		  \dimen3=\sine\dimen1
		  \global\advance\p@intvaluex by -\dimen3
		  \global\p@intvaluey=\sine\dimen0
		  \dimen3=\cosine\dimen1
		  \global\advance\p@intvaluey by \dimen3
		  }}
\def\compute@bb{
		\no@bbfalse
		\if@bbllx \else \no@bbtrue \fi
		\if@bblly \else \no@bbtrue \fi
		\if@bburx \else \no@bbtrue \fi
		\if@bbury \else \no@bbtrue \fi
		\ifno@bb \bb@missing \fi
		\ifno@bb \ps@typeout{FATAL ERROR: no bb supplied or found}
			\no-bb-error
		\fi
		\count203=\@p@sbburx
		\count204=\@p@sbbury
		\advance\count203 by -\@p@sbbllx
		\advance\count204 by -\@p@sbblly
		\edef\ps@bbw{\number\count203}
		\edef\ps@bbh{\number\count204}
		\if@angle 
			\Sine{\@p@sangle}\Cosine{\@p@sangle}
	        	{\dimen100=\maxdimen\xdef\r@p@sbbllx{\number\dimen100}
					    \xdef\r@p@sbblly{\number\dimen100}
			                    \xdef\r@p@sbburx{-\number\dimen100}
					    \xdef\r@p@sbbury{-\number\dimen100}}
                        \def\minmaxtest{
			   \ifnum\number\p@intvaluex<\r@p@sbbllx
			      \xdef\r@p@sbbllx{\number\p@intvaluex}\fi
			   \ifnum\number\p@intvaluex>\r@p@sbburx
			      \xdef\r@p@sbburx{\number\p@intvaluex}\fi
			   \ifnum\number\p@intvaluey<\r@p@sbblly
			      \xdef\r@p@sbblly{\number\p@intvaluey}\fi
			   \ifnum\number\p@intvaluey>\r@p@sbbury
			      \xdef\r@p@sbbury{\number\p@intvaluey}\fi
			   }
			\rotate@{\@p@sbbllx}{\@p@sbblly}
			\minmaxtest
			\rotate@{\@p@sbbllx}{\@p@sbbury}
			\minmaxtest
			\rotate@{\@p@sbburx}{\@p@sbblly}
			\minmaxtest
			\rotate@{\@p@sbburx}{\@p@sbbury}
			\minmaxtest
			\edef\@p@sbbllx{\r@p@sbbllx}\edef\@p@sbblly{\r@p@sbblly}
			\edef\@p@sbburx{\r@p@sbburx}\edef\@p@sbbury{\r@p@sbbury}
		\fi
		\count203=\@p@sbburx
		\count204=\@p@sbbury
		\advance\count203 by -\@p@sbbllx
		\advance\count204 by -\@p@sbblly
		\edef\@bbw{\number\count203}
		\edef\@bbh{\number\count204}
}
\def\in@hundreds#1#2#3{\count240=#2 \count241=#3
		     \count100=\count240	% 100 is first digit #2/#3
		     \divide\count100 by \count241
		     \count101=\count100
		     \multiply\count101 by \count241
		     \advance\count240 by -\count101
		     \multiply\count240 by 10
		     \count101=\count240	%101 is second digit of #2/#3
		     \divide\count101 by \count241
		     \count102=\count101
		     \multiply\count102 by \count241
		     \advance\count240 by -\count102
		     \multiply\count240 by 10
		     \count102=\count240	% 102 is the third digit
		     \divide\count102 by \count241
		     \count200=#1\count205=0
		     \count201=\count200
			\multiply\count201 by \count100
		 	\advance\count205 by \count201
		     \count201=\count200
			\divide\count201 by 10
			\multiply\count201 by \count101
			\advance\count205 by \count201
		     \count201=\count200
			\divide\count201 by 100
			\multiply\count201 by \count102
			\advance\count205 by \count201
		     \edef\@result{\number\count205}
}
\def\compute@wfromh{
		\in@hundreds{\@p@sheight}{\@bbw}{\@bbh}
		\edef\@p@swidth{\@result}
}
\def\compute@hfromw{
	        \in@hundreds{\@p@swidth}{\@bbh}{\@bbw}
		\edef\@p@sheight{\@result}
}
\def\compute@handw{
		\if@height 
			\if@width
			\else
				\compute@wfromh
			\fi
		\else 
			\if@width
				\compute@hfromw
			\else
				\edef\@p@sheight{\@bbh}
				\edef\@p@swidth{\@bbw}
			\fi
		\fi
}
\def\compute@resv{
		\if@rheight \else \edef\@p@srheight{\@p@sheight} \fi
		\if@rwidth \else \edef\@p@srwidth{\@p@swidth} \fi
}
\def\compute@sizes{
	\compute@bb
	\if@scalefirst\if@angle
	\if@width
	   \in@hundreds{\@p@swidth}{\@bbw}{\ps@bbw}
	   \edef\@p@swidth{\@result}
	\fi
	\if@height
	   \in@hundreds{\@p@sheight}{\@bbh}{\ps@bbh}
	   \edef\@p@sheight{\@result}
	\fi
	\fi\fi
	\compute@handw
	\compute@resv}

\def\psfig#1{\vbox {
	\ps@init@parms
	\parse@ps@parms{#1}
	\compute@sizes
	\ifnum\@p@scost<\@psdraft{
		\special{ps::[begin] 	\@p@swidth \space \@p@sheight \space
				\@p@sbbllx \space \@p@sbblly \space
				\@p@sbburx \space \@p@sbbury \space
				startTexFig \space }
		\if@angle
			\special {ps:: \@p@sangle \space rotate \space} 
		\fi
		\if@clip{
			\if@verbose{
				\ps@typeout{(clip)}
			}\fi
			\special{ps:: doclip \space }
		}\fi
		\if@prologfile
		    \special{ps: plotfile \@prologfileval \space } \fi
		\if@decmpr{
			\if@verbose{
				\ps@typeout{psfig: including \@p@sfile.Z \space }
			}\fi
			\special{ps: plotfile "`zcat \@p@sfile.Z" \space }
		}\else{
			\if@verbose{
				\ps@typeout{psfig: including \@p@sfile \space }
			}\fi
			\special{ps: plotfile \@p@sfile \space }
		}\fi
		\if@postlogfile
		    \special{ps: plotfile \@postlogfileval \space } \fi
		\special{ps::[end] endTexFig \space }
		\vbox to \@p@srheight sp{
			\hbox to \@p@srwidth sp{
				\hss
			}
		\vss
		}
	}\else{
		\if@draftbox{		
			\hbox{\frame{\vbox to \@p@srheight sp{
			\vss
			\hbox to \@p@srwidth sp{ \hss \@p@sfile \hss }
			\vss
			}}}
		}\else{
			\vbox to \@p@srheight sp{
			\vss
			\hbox to \@p@srwidth sp{\hss}
			\vss
			}
		}\fi

	}\fi
}}
\psfigRestoreAt
\let\@=\LaTeXAtSign

\includeonly {}
\setlength{\topmargin}{0.0in}
\setlength{\oddsidemargin}{0.0in}
\setlength{\footskip}{0.6in}
\setlength{\headheight}{0.0in}
\setlength{\headsep}{0.0in}
\setlength{\textheight}{9.0in}
\setlength{\textwidth}{6.7in}
\setlength{\parindent}{0in}
\setlength{\parskip}{0.1in}
\setlength{\columnsep}{2.5pc}

\renewcommand{\baselinestretch}{1.0}

\begin {document}

\title{A Novel Representation to Improve Team Problem Solving in Real-Time}
\author{ Alex Doboli\\
 Department of Electrical and Computer Engineering\\  Stony Brook University, Stony Brook NY 11794-2350\\ Email: alex.doboli@stonybrook.edu 
}
\date{}
\maketitle

\begin{abstract}

This paper proposes a novel representation to support computing metrics that help understanding and improving in real-time a team's behavior during problem solving in real-life. 
Even though teams are important in modern activities, there is little computing aid to improve their activity. The representation captures the different mental images developed, enhanced, and utilized during solving. A case study illustrates the representation.    

\end{abstract}

%\addtolength{\textwidth}{0.5in} \addtolength{\textheight}{0.8in}
%\addtolength{\topmargin}{-0.2in}
%\addtolength{\oddsidemargin}{-0.2in}
%\addtolength{\evensidemargin}{-0.2in}

%\renewcommand{\baselinestretch}{0.958}
%\documentclass[journal]{}

\thispagestyle{empty}

\section {Introduction}

Teams are essential in the majority of activities and endeavors due to their increasing complexity and cross-disciplinary nature~\cite{Fiore2010, Salas2008, Graesser2018}. It is unlikely that single individuals working in isolation can address these challenges~\cite{Hong2004}. Research in psychology, sociology and organization science has uncovered many important factors that condition team effectiveness, like coordination, synchronicity and divergence of ideas, similarity of goals, motivation, trust, and so on~\cite{Salas2008}. Team behavior has been often studied in laboratory conditions using recorded verbal communications between team members and questionnaires submitted at the end of a an experimental study~\cite{Orasanu1992}. However, there is limited work on understanding team behavior during actual problem solving in real-life conditions and how to improve individual's behavior to improve a team's results. A main reason is the absence of adequate computing support.

This paper presents a novel representation to support computing in real-time metrics describing team behavior during problem solving in real-life conditions. The representation captures eight different mental images developed, enhanced, and utilized during problem solving, e.g., images of problem requirements, desired and existing solutions, expected and observed behavior, causality of the differences between expectation and observation, and needed changes to a solution and problem requirements. The representation is based on Idea Cluster Nodes (ICNs) that group similar ideas. ICNs are related to each other to express detailing, exploration, causality, and generalization. A case study illustrates the proposed representation.    

The paper has the following structure. Section II presents an example of team problem solving. Section III details the problem-solving model. Model requirements are stated in Section IV. Section V indicates the proposed representation. Conclusions end the paper. 

\section {Example of Team Problem Solving}

The following exercise describes the type of problems that considered in our work. The problem requires reading data from an input file. The data includes date, temperature, humidity, and air pressure. Multiple readings exist for the same date. Each read value must be in a predefined range, but the input values can be incorrect. The problem requires to find the dates with the most incorrect input values. In addition, the problem description includes a set of typical input data and the associated outputs. This data adds clarity about any possible uncertainties about the problem description. Still, uncertainties and unknowns about the problem description might still exist, like if all readings for the same date are in sequence or mixed up with other dates. 
%It can be argued that many problem descriptions remain ambiguous 
%or undefined despite efforts to clarify all their details. 

The team activity during problem solving can be observed over time to characterize the emerging team dynamic during problem solving~\cite{Doboli2021,Duke2022a,Duke2022b,Liu2015}. For example, using the system in~\cite{Duke2022a}, the following discussion trace of a team problem solving situation can be produced. Each input ${\bf I_i}$ is a different idea suggested by one of the team members. 
{\small ${\bf I_1}$: ``{\tt Compare to find a larger sum}''. ${\bf I_2}$: ``{\tt Add up all numbers and see the larger}''. ${\bf I_3}$: ``{\tt Brute-force, add-up all combinations}''. ${\bf I_4}$: ``{\tt Many loops}''. ${\bf I_5}$: ``{\tt Brute-force, take the length of entire and check sum}''. ${\bf I_6}$: ``{\tt Consider length minus one}''. ${\bf I_7}$: ``{\tt Shift left}''. ${\bf I_8}$: ``{\tt Check how many combinations}''. ${\bf I_9}$: ``{\tt Check length}''. ${\bf I_{10}}$: ``{\tt Subtract length}''. ${\bf I_{11}}$: ``{\tt Smack in the middle}''. ${\bf I_{12}}$: ``{\tt Length is four, index must change}''. ${\bf I_{13}}$: ``{\tt Change a lot of things}''. ${\bf I_{14}}$: ``{\tt Consider edge, the other side}''. ${\bf I_{15}}$: ``{\tt That would work. A lot of brute-force}''. ${\bf I_{16}}$: ``{\tt Bubble sort, find the largest sum}''. ${\bf I_{17}}$: ``{\tt Bubble sort might work}''. ${\bf I_{18}}$: ``{\tt Like truth tables, each combination in a row, add them up}''. ${\bf I_{19}}$: ``{\tt Array has at most thirty values. A lot of brute force, 360 loops}''}.

The next section discusses the procedure to model a team problem-solving process.

\begin{figure}\centering
\includegraphics[width=3.4in]{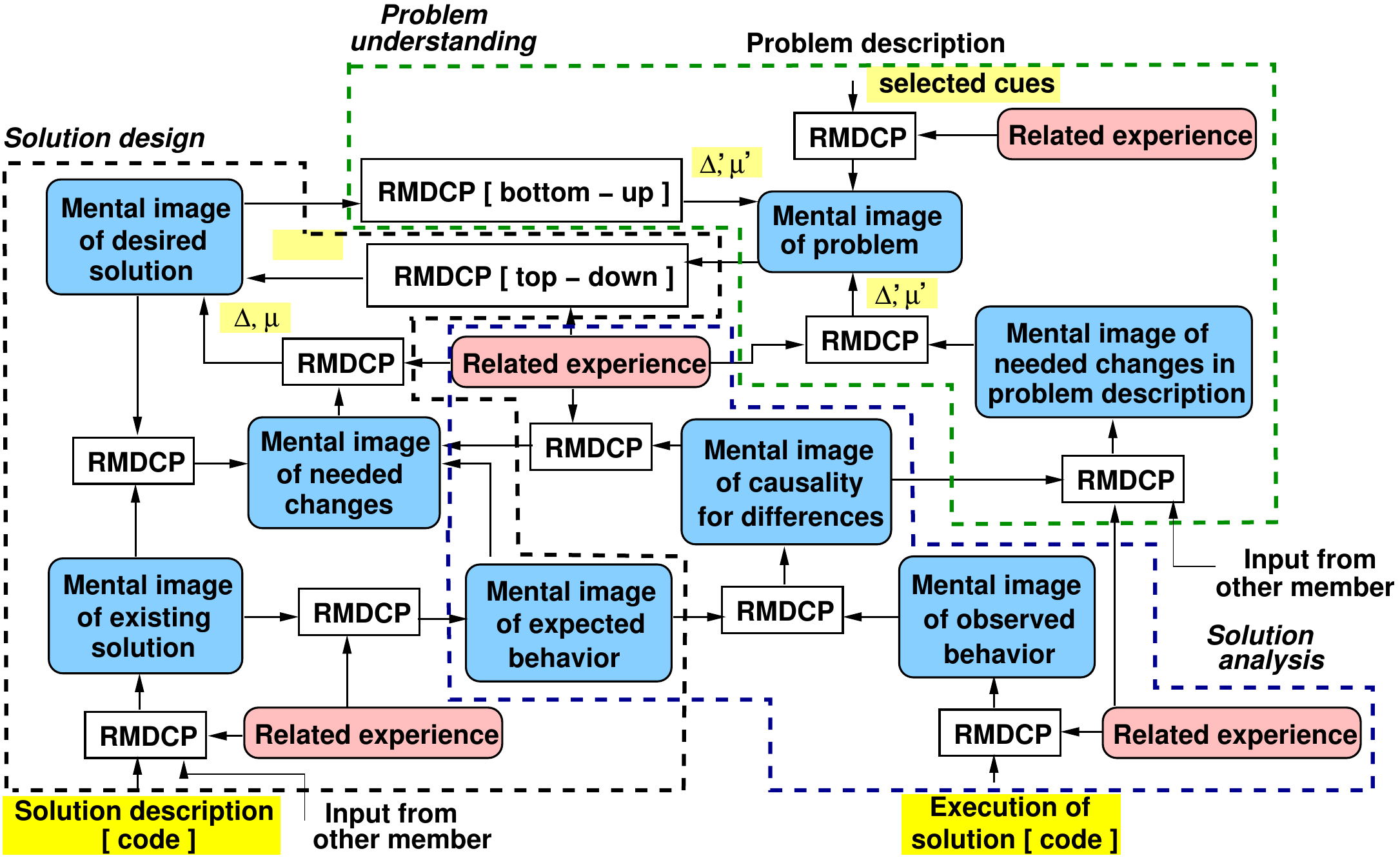}
\caption{Conceptual model for problem understanding and solving}
\label{SAUCE_0}
\vspace * {-0.2in}
\end{figure}

\section {Modeling Individual Problem Solving}

Problem descriptions include functional (processing) and data requirements, including output values that are expected to result for specific inputs. Unknowns and ambiguities can be part of a problem description. High-level solution descriptions are mixtures of four kinds of descriptions: (a)~a sequence of actions (e.g., processing), sometimes expressed as pseudocode, (b)~a set of goals that must be achieved, (c)~a set of requirements for the outputs (i.e. logic conditions that they should meet), and (d)~a set of inputs and corresponding outputs expected for the solutions. It is unclear what cognitive, social or emotional parameters influence the mixture of the four feature types. The purpose of the problem-solving process is to devise a solution that minimizes the difference between the requirements of the problem description and the features of a devised solution.

The model for problem solving by an individual includes the multiple images used in solving (e.g., work memory), the nature of knowledge representation (i.e. long-term memory), the individual's goals and expectation about being able to solve the problem and the resulting outcomes, and the sequence of activities over time as part of problem solving. Figure~\ref{SAUCE_0} illustrates the proposed conceptual model for an individual's problem understanding and solving activities. The flow considers two separate mental images, an image of the problem description and one of the solutions. Both images can be mixtures of the above features (a)-(d). The problem image results from the problem understanding activity. The solution image is created by solution analysis. Changes of the problem requirements images during problem solving are described by pair $\Delta', \mu'$, where the first term indicates the change of an image and the second represents the meaning of the change. The changes of the solution images are shown as $\Delta, \mu$. A solution is successfully created when $\Delta, \mu \approx \Delta', \mu' \approx 0$. 

The difference between two images is established by RMDCP (Recall - Match - Difference - Combine - Predict) process, which includes four steps: recalling the two images from the memory, matching the images to relate their similar parts, finding the difference of the two, combining the differences of a new image to the image recalled from the memory, and predicting the meaning of the difference in the context of the common matched parts. A bottom-up RMDCP process starting from the image of an existing solution identifies the difference $\Delta',\mu'$ of the problem image, and a top-down RMDCP process starting from the current image of the problem finds the difference $\Delta, \mu$ of the solution image. RMDCP process is executed in the context of the related experience (i.e. existing knowledge) of the member.

A main feature of the model in Figure~\ref{SAUCE_0} is the using of two images connected by the two RMDCPs. Having two images offers the following benefits: (i)~The image of the problem can serve as a reference to set goals for problem solving. It has invariant requirements (even though ambiguities and unspecified elements can be included). Moreover, the image should be abstract enough to support exploration during problem solving. (ii)~The matching and difference computation between the two images serves to guide the decision making during problem solving, such as by $\Delta, \mu \approx \Delta', \mu' \approx 0$, or by maximizing the distance $\Delta, \mu \approx \Delta', \mu'$~\cite{Duke2022b}. (iii)~Matching, difference, and prediction between the two images supports decomposing goals into sub-goals to decide what should be solved next, and also to predict causality, such as how different solution parts contribute to the achieving the problem requirements. Goal decomposition and causality understanding support setting priorities among sub-goals and causal relations, hence guiding decision making during solving process. It also allows to define the difference between the expected and the observed effect of a solution fragment. Finally, the two RMDCPs can express the following two situations: (a)~If they continuously relate the two images then they mimic the subconscious activity that finally creates the feeling that something is right or not, or possibly express sudden insight. (b)~They can represent the conscious activity when a member directs his / her attention on finding the differences between the current solution and problem requirements. 

The description of the resulting differences $\Delta, \mu$ of the top-down RMDCP (which relate the solution to the problem requirements) becomes the current sub-goal of solution design. The solution is then executed. The solution and the execution behavior (like the outputs generated for specific inputs) are inputs to the solution analysis activity. Solution analysis determines differences $\Delta, \mu$ between the expected and observed execution behavior. The differences represent missing or incorrect parts of the solution. The differences modify the mental images of the solution. The analysis can involve reasoning based on the code, such as mentally executing the code. Alternatively, analysis can be based on the outputs obtained by executing the code for specific inputs. Reasoning attempts to identify the required change $\Delta, \mu$ to the solution image by following backwards the causal sequence that created the unwanted result. Data-based analysis obtains changes $\Delta, \mu$ by generalizing the input - output behavior  as a rule that creates this behavior, and then combining it with the solution image. Hence, the mental images of a solution, solution design and execution, and analysis form a bottom-up flow that can involve backwards reasoning about the processing steps along the solution's causal relations and using the differences between the expected and observed inputs and outputs to create the processing steps of the solution.  

Hence, as shown in Figure~\ref{SAUCE_0}, the mental images included in the model for an agent's problem solving process are as follows: (i)~mental image of the problem, (ii)~mental image of the desired solution, (iii)~mental image of the existing solution, (iv)~mental image of the expected behavior, (v)~mental image of the observed behavior, (vi)~mental image of the causality of the differences between the expected and observed behaviors, (vii)~mental images of the needed changes, and (viii)~mental images of the needed changes for the problem description. The eight images are shown as blue boxes. Note that the eight mental images are not directly observable, but only understandable based on an agent's outputs during problem solving, like produced pseudocode and code, verbal communications, and social, and emotional reactions. The next section presents the foundations for the proposed model to represent the eight images: metrics and systematic analysis of team problem solving. 

\vspace * {-0.1in}
\section {Requirements for Team Problem Solving Analysis: Metrics and systematic analysis}
\vspace * {-0.1in}

\subsection {Metrics}

The following broad categories of metrics describe an individual's or a team's problem solving behavior:

(1) {\em Fulfilled requirements}: Metrics present the number of requirements, observed behavior, implementation fragments, and variables that were correctly and fully addressed by the problem solving process. They characterize the robustness of the solving process and solutions. 

(2) {\em Exploration}: Metrics refer to the capacity to switch between ideas, including different solution approaches and detailed implementation ideas. They characterize the degree to which possible alternatives were identified and analyzed to establish their pros and cons.     

(3) {\em Substantiated decisions}: Metrics quantify the degree to which solution decisions are motivated (explained) by the previous decisions, solution context, and experience.

\begin{figure}\centering
\includegraphics[width=3.2in]{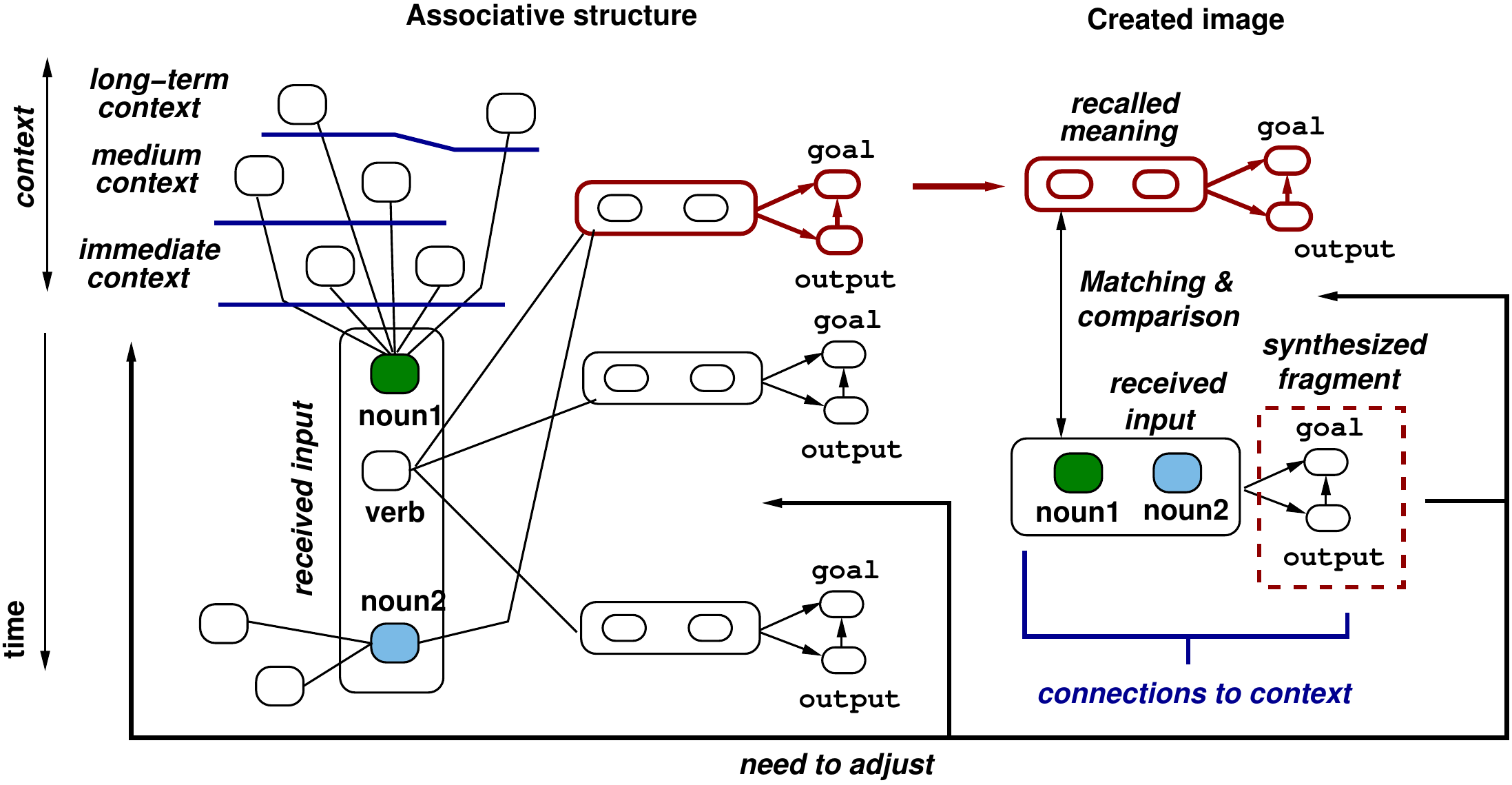}
\caption{Principle for creating the image of a problem solving input}
\label{SAUCE_7}
\vspace * {-0.2in}
\end{figure}

(4) {\em Backtracking}: Metrics indicate the capability to identify the cause of errors or the places to be changed for a certain motivation, like unaddressed requirements or unwanted observed behavior. They suggest the prioritization of goals and the degree to which the causal dependencies of a solution are correctly understood. 

(5) {\em Contradictions}: Metrics describe the nature and number of conflicts, contradictions, or unaddressed (e.g., remained unspecified) between the requirements and assumptions of the different images, like problem requirements and desired solution, needed changes and desired or existing solutions, objectives (including personal objectives) and individual experience, and so on. They indicate possible flaws that are embedded into all solution implementations.

(6) {\em Repetitions}: Metrics express the nature and number of ideas that are repeated during the solving process, as well as the results of these repetitions, including new ideas and addressed inconsistencies and errors. Repetitions can point towards getting additional insight into a solution, such as when errors are corrected or improvements are suggested, or towards a situation when an individual or team reached a situation of design fixation, like it is unclear how to continue.    

(7) {\em Unconsidered needs}: Metrics capture the unaddressed problem requirements, solution fragments or observed behavior that were not analyzed, and unconsidered necessary changes in a solution or problem understanding. They describe liabilities of the solving process as they express uncovered elements of a solution. 

(8) {\em Unexplored items}: Metrics refer to possibilities that were not analyzed during problem solving, such as different algorithms or processing flows for a problem. They present potential inefficiencies of a solution.      

\begin{figure*}\centering
\includegraphics[width=6.0in]{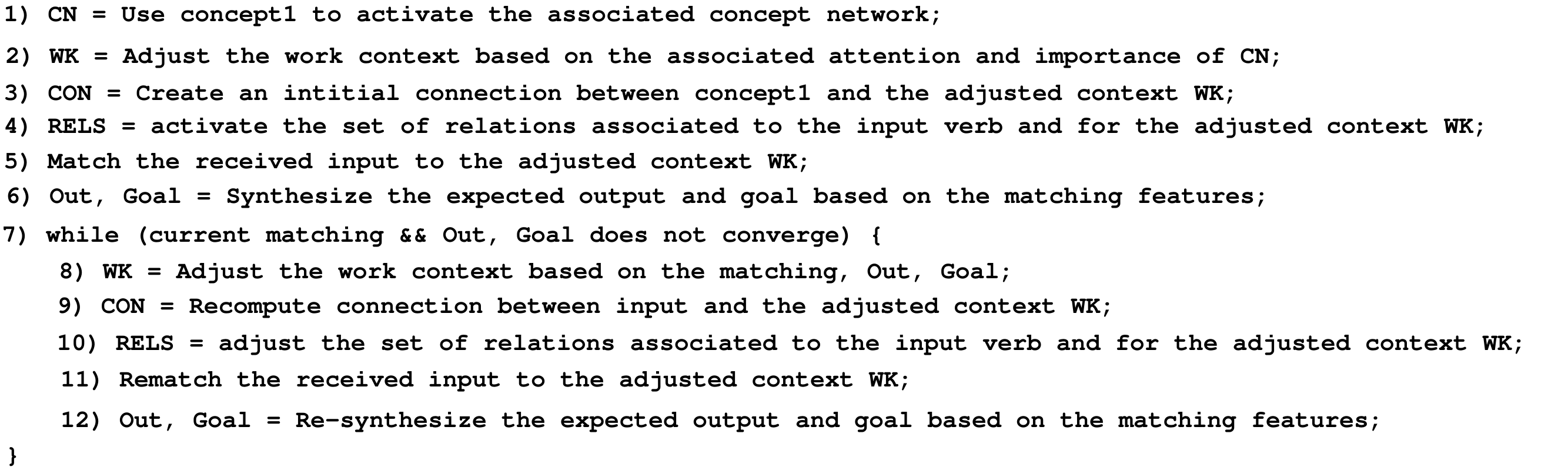}
\caption{Algorithm for creating the image of the problem solving process}
\label{SAUCE_6}
\vspace * {-0.2in}
\end{figure*}

\vspace * {-0.1in}
\subsection {Systematic Analysis of Team Problem Solving} 
\vspace * {-0.1in}

Figure~\ref{SAUCE_7} depicts the pursued principle to create the image of a new input for solving a problem, and Figure~\ref{SAUCE_6} shows the algorithm to construct the image of problem solving activity in a team. The received input in Figure~\ref{SAUCE_7} is the sequence of $noun_1$ (green block), verb, and $noun_2$ (blue block), which must be understood and connected to the previous inputs as part of image creation. As $noun_1$ is the first part of the new input received over time, it is used to activate the context in which the meaning of the input is synthesized (understood). Instructions (1) - (3) in Figure~\ref{SAUCE_6} describe the steps. The context is qualitatively partitioned into immediate context, which includes the previous inputs in a small time window, the medium context to store the main fragments of the solution, and the long-term context to keep the knowledge available for problem solving. The verb of the input activates the associated relations (meanings) of the verb, and then $noun_2$ selects the current relation from the activated set (instruction (4) in Figure~\ref{SAUCE_6}). The recalled relation is shown in red in the figure. The relation might include an associated output, goal, and degree to which the output matches the goal. The similarity and dissimilarity between the received input and recalled meaning is found through matching and comparison of the two (instruction (5) in Figure~\ref{SAUCE_6}). Matching finds the meaning-wise similar parts of the two, and comparison decides their commonalities and differences. Then, understanding the input requires synthesizing the expected goal and output of the input based on the matching and comparison results as well as producing all other connections to the context (instruction (6) in Figure~\ref{SAUCE_6}). It is possible that the synthesized meaning and connections of the input to the context produce a ``feeling'' of incorrect, and thus, produce the need to adjust (instruction (7) in Figure~\ref{SAUCE_6}). Adjusting can be in three places: the matching of the input to the recalled meaning, the associated meaning of the verb (i.e. relation), and the context in which input is understood. Instructions (8) - (12) describe adjusting.     

\vspace * {-0.1in}
\subsection {Summary of the Involved Team Solving Activities and Characteristics}
\vspace * {-0.1in}

The following aspects must be captured during the analysis and modeling of problem solving in teams:

(1) Decide the nature of an input: It includes finding if an input pertains to problem understanding or solving. Inputs can be clustered based on whether they relate to problem understanding or solving, but some inputs of a kind might be included among the inputs pertaining to the other purpose. For example, a high-level idea of solving the problem was embedded in the questions asked to clarify the problem description. For inputs related to solving, the activity also identifies if an input represents explaining or understanding, such as high-level solution description or detailing, comparison, analysis to understand the pros and cons of a solution, finding missing fragments (which can become sub-goals for problem solving), localization of solving, identification of required changes, and combining previous ideas.

(2) Finding the connection between two inputs: It involves identifying the previous inputs as well as the specific input fragments to which the current input relates. Ambiguities (unknowns) on correct connection identification might remain. Some statements represent broad statements, including approvals, truisms, or other broad generalizations, which are hard to connect to a specific input. The activity refers to ideas on both problem understanding and solving. Also, it is possible that the connection is to the entire (previous) input, some of its fragments, or certain unspecified but available (thus implicit) features related to the idea and its fragments. Moreover, the current input can connect to the strictly previous idea and to the previous ideas within a short time window (e.g., working memory), the main ideas within a time window of average length, or to the permanent ideas that pertain to the long-term memory. A certain idea connection can evolve during problem solving, as new nuances of idea relatedness are uncovered, such as new features that are linked or new relations that are established. Previously unrelated ideas can become connected too.  
 
(3) Understanding the context of an input: The meanings of the received inputs depend on their connections with contexts of various characteristics. A context can have various degrees of coverings, such as an immediate covering set by the previous inputs and that spans over a small number of next inputs, or a broad covering defined by accepted facts stored in the long-term memory. It is important to determine how the nature and sequence of the received inputs influences the degrees of the contextual coverings, e.g., the influence of specific nouns, verbs, repetitions, restatements, and so on. Also, understanding refers to the mechanism of including some structures into the context because they are understood and accepted by a team, while other concepts, even though potentially significant, are discarded.     

(4) Finding the maximum matching between two inputs: Identify missing (implicit) fragments that could be reasonably justified by the already matched fragments and/or connection with previous inputs. 

(5) Identifying semantically related ideas of different structures: It refers to ideas that express similar meanings, e.g., outputs or effects, through inputs that utilize different nouns and actions (verbs) and/or are structurally different. It also includes situations when inputs have opposite meanings, such as referring to a $gap$ in the context of explaining the meaning of a $contiguous~subsequence$. 

(6) Finding the level of abstraction of inputs, including ideas on problem understanding and solutions. 

(7) Identifying fragments of previous inputs to be included in future inputs, including solutions: It includes fragments that are utilized in an opposite way as their purpose in previous inputs (context). 

(8) Associating the code implementation to the discussed ideas.

(9) Creating the overall image (map) of all proposed high-level solution descriptions: The stated high-level descriptions describe the spanned solution space, and if certain solution opportunities are missed. It is important to characterize the similarity and dissimilarity of the description using description matching and identifying unmatched or missing fragments in the descriptions. 
%As summarized in Figure~\ref{SAUCE_4}, 
Some description express the same meaning on how the problem should be solved (even though using different structures), but other descriptions offer fundamentally different ideas, such as idea ${\bf I_{16}}$ and its subsequent elaboration ${\bf I_{18}}$ as compared to the other high-level solution ideas. 

(10) Relating the overall image or individual high-level descriptions to the correct solution idea: Identify missing (implicit) fragments.

\begin{figure}\centering
\includegraphics[width=3.4in]{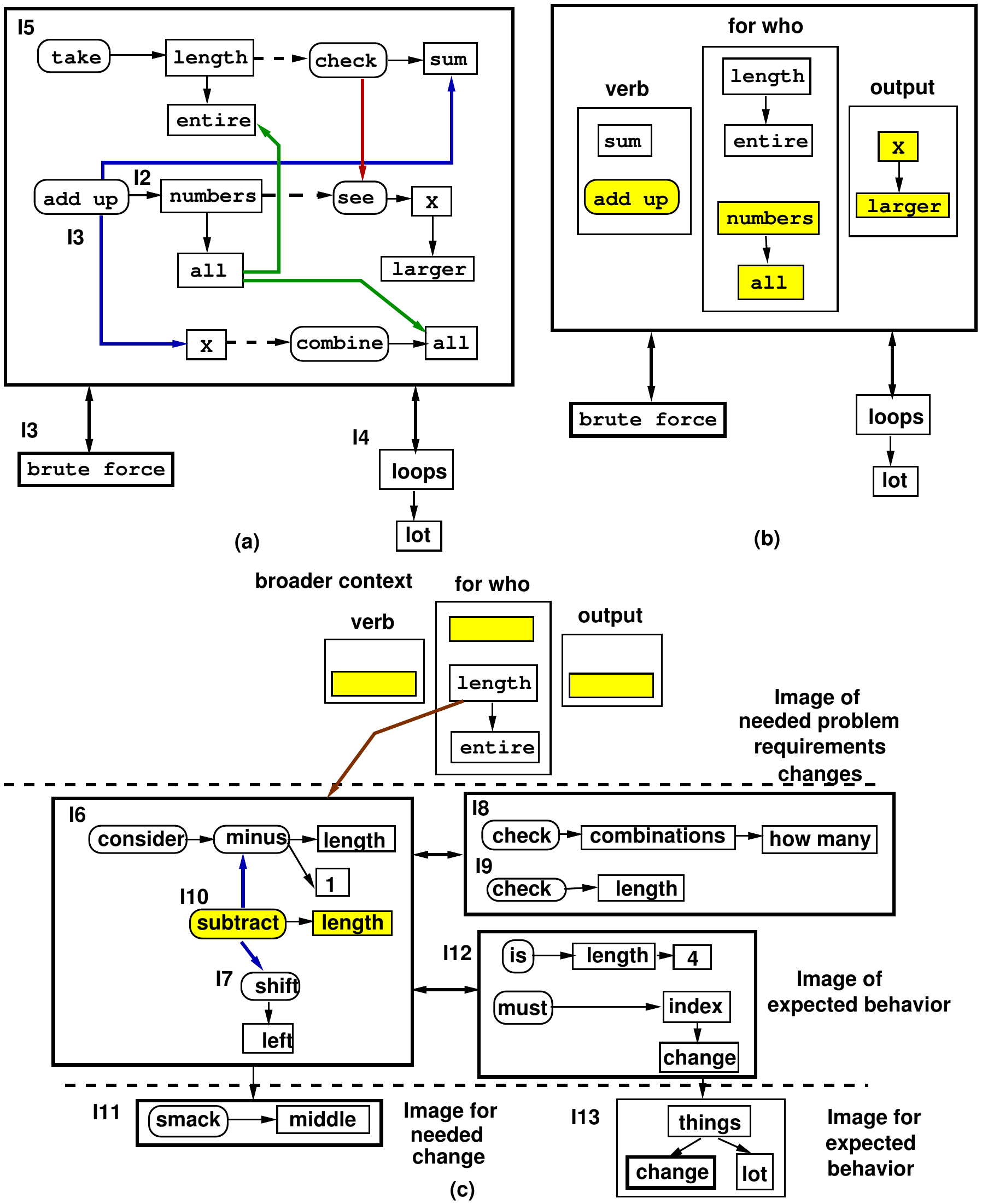}
\caption{Representation to express mental image evolution during problem solving: (a) related solution ideas, (b) Idea Cluster Node, and (c) detailing and its related Idea Cluster Nodes}
\label{SAUCE_9}
\vspace * {-0.1in}
\end{figure}

\section {Proposed Representation}

Figure~\ref{SAUCE_9} illustrates the main elements of the proposed representation to express the mental image evolution of agents and teams during problem solving.

The basic element of the representation is the {\em Ideas Cluster Node} (ICN) shown in Figure~\ref{SAUCE_9}(b) for the ideas in Figure~\ref{SAUCE_9}(a). An ICN describes all similar ideas in a solution. Each idea includes a verb and two nouns as depicted in Figure~\ref{SAUCE_7}. Each new idea is related to the image recalled from the memory through matching and comparison of the verbs and nouns, as in Figure~\ref{SAUCE_7}. For example, Figure~\ref{SAUCE_7}(a) presents the relation between ideas ${\bf I_2}$, ${\bf I_3}$, ${\bf I_4}$, and ${\bf I_5}$ of the example in Section~II. The green, blue, and red arrows indicate the matched elements. Also, the figure includes ``{\tt brute force}'' pertaining to the image about the desired solution, and ythe idea ``{\tt lot of loops}'' expressing an expected feature of the solution behavior. The corresponding ICN in Figure~\ref{SAUCE_9}(b) shows the {\em typical elements} (TE) (highlighted in yellow) of the related ideas as well as the {\em expected variation} (EV) from the typical elements. An ICN includes three specific slots depending on the nature of the ideas: for example, ideas that express processing have one slot for the verbs, one for the nouns on which the action is performed (for who), and one for the expected outputs. Note that Figure~\ref{SAUCE_9} exemplifies different images that occur during problem solving.

{\em Detailing} includes ICNs that describe the meaning or implementation of a group of elements of another ICN. For example, the detailing shown in Figure~\ref{SAUCE_9}(c) presents the ICN corresponding to inputs ${\bf I_6}$, ${\bf I_7}$, and ${\bf I_{10}}$, and which details the elements ``{\tt entire length}'' of the ICN in Figure~\ref{SAUCE_9}(b). The detailing ICN represents an additional step during problem solving, such as it focuses on further devising a specific fragment within the broader context set by the global ICN in Figure~\ref{SAUCE_9}(b). Hence, the detailing ICN must remain valid within the context of the global ICN. For the detailing ICN, idea ${\bf I_{12}}$ represents an image of the expected behavior, and ideas ${\bf I_8}$ and ${\bf I_9}$ pertain to the image of the needed changes of the problem requirements. ICN for idea ${\bf I_{11}}$ indicates an image of the needed solution changes, and ICN for idea ${\bf I_{13}}$ is part of the expected behavior image.   

\begin{figure}\centering
\includegraphics[width=3.2in]{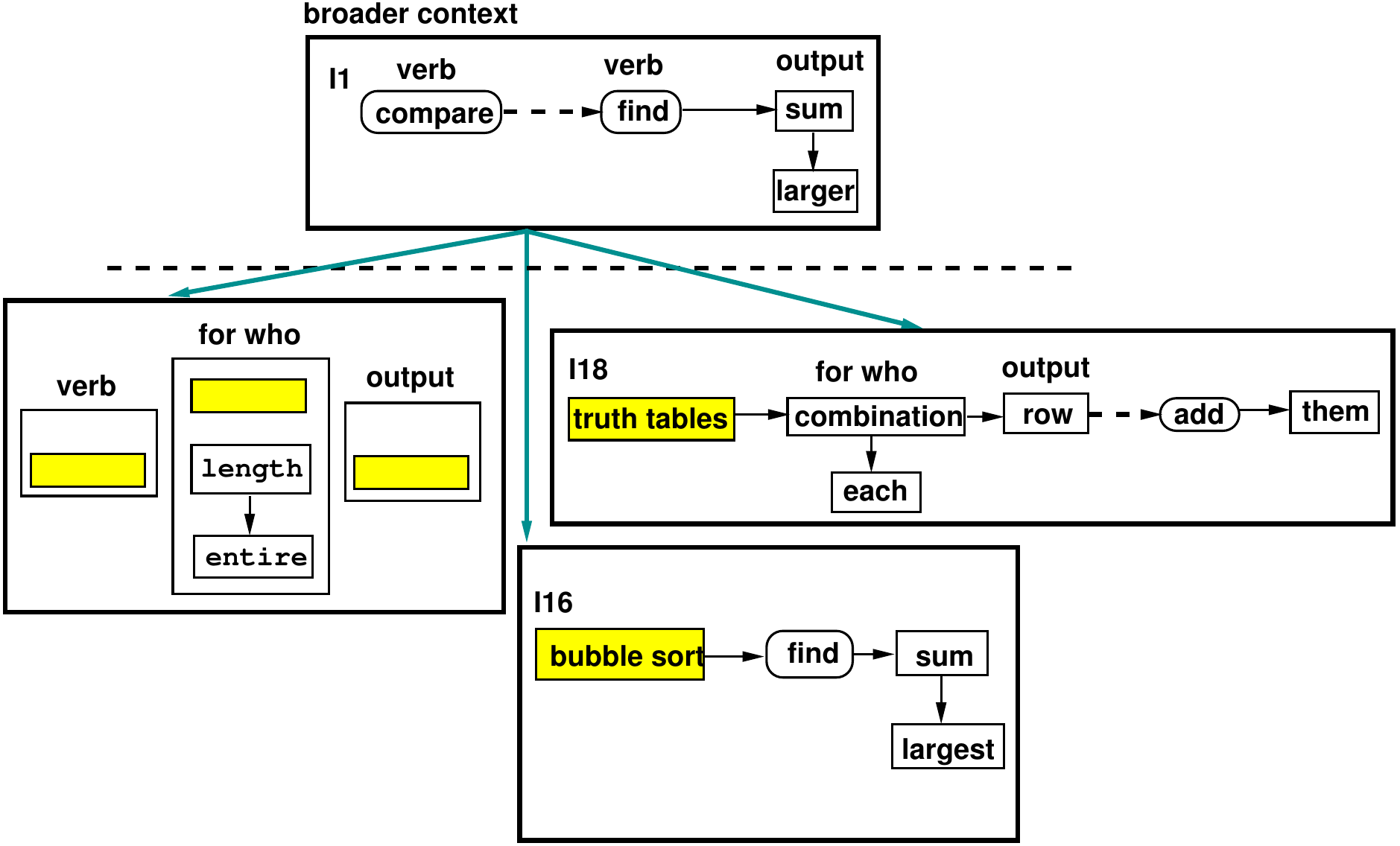}
\caption{Representation of the mental images during exploration}
\label{SAUCE_10}
\vspace * {-0.2in}
\end{figure}

{\em Exploration} introduces ICNs that alternative ideas for an ICN setting a broader context. For example, Figure~\ref{SAUCE_10} shows three alternative ICNs for the ICN of the starting idea ${\bf I_1}$. The ICN for idea ${\bf I_{16}}$ is an alternative to the ideas expressed by the ICN in Figure~\ref{SAUCE_9}(b). An other alternative is the ICN for idea ${\bf I_{18}}$.

\begin{figure}\centering
\includegraphics[width=1.6in]{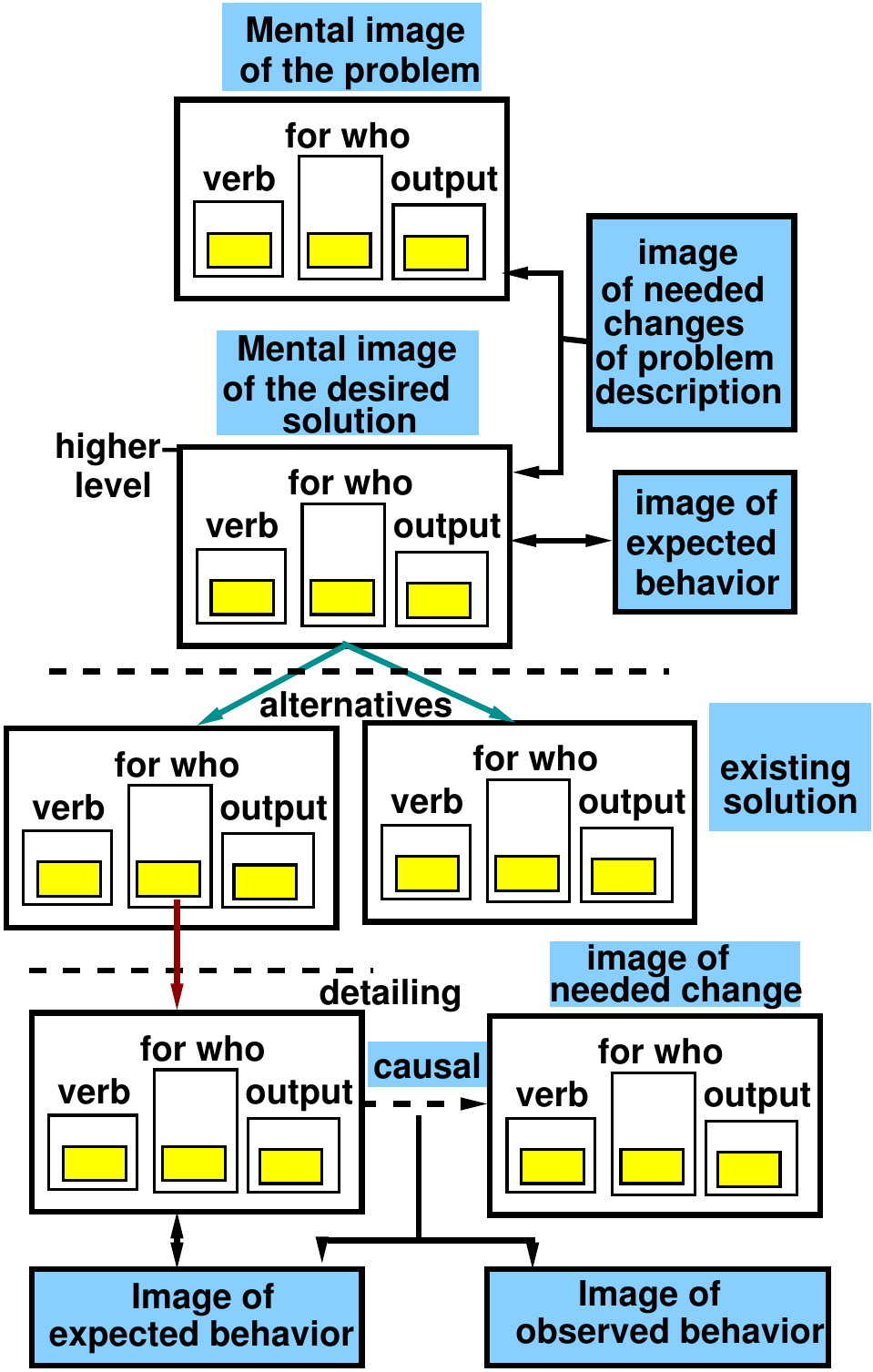}
\caption{Representation for problem solving process}
\label{SAUCE_11}
\vspace * {-0.2in}
\end{figure}

Figure~\ref{SAUCE_11} depicts the proposed representation expressing a problem solving process. It includes the previously mentioned constructs.

\vspace * {-0.1in}
\section {Conclusions}
\vspace * {-0.1in}

This paper presents a novel representation to support computing in real-time metrics to improve team behavior during problem solving in real-life conditions. The representation captures the different mental images developed, enhanced, and employed during problem solving. A case study illustrates the proposed representation. Future work will use the representation to devise computational tools for team behavior enhancing.

\bibliographystyle{unsrt}
%\bibliography {wp}

\end {document}